\documentclass[letterpaper, 10 pt, conference]{ieeeconf}  

\IEEEoverridecommandlockouts

\usepackage{amsmath,amssymb,amsfonts}
\usepackage{graphicx}
\usepackage{cite}
\usepackage{algorithm}
\usepackage{algorithmic}
\usepackage{subcaption}
\usepackage{float} 

\usepackage{xcolor}
\usepackage{soul}
\usepackage{hyperref}

\hypersetup{
    colorlinks = true,
    linkcolor = blue,
    anchorcolor = red,
    citecolor = magenta,
    filecolor = cyan,
    menucolor = red,
    runcolor = cyan,
    urlcolor = magenta
}

\begin{document}

\title{SHIFT Planner: Speedy Hybrid Iterative Field and Segmented Trajectory Optimization with IKD-tree for Uniform Lightweight Coverage}

\author{
    Zexuan Fan\textsuperscript{1,†},
    Sunchun Zhou\textsuperscript{1},
    Hengye Yang\textsuperscript{1},
    Junyi Cai\textsuperscript{1},
    Ran Cheng\textsuperscript{1},
    Lige Liu\textsuperscript{1},
    Tao Sun\textsuperscript{1,†}
    \thanks{
        \textsuperscript{1} Robotics Algorithm Team, Microwave $\&$ Cleaning Appliance Division, Midea Group, China.
    }
    \thanks{
        † Corresponding Authors
    }
    \thanks{
        Emails:\texttt{\{zexuan.fan, zhousc18, yanghy228, caijy20, chengran1, liulg12, tsun\}@midea.com}
    }
    \thanks{
        Open-source implementation is available at \url{https://fanzexuan.github.io/SHIFTPlanner/}.
    }
}

\maketitle

\begin{abstract}
Achieving uniform coverage in dynamic environments is essential for autonomous robots in cleaning, inspection, and agricultural operations. 
Unlike most existing approaches that prioritize path length and time optimality, we propose the SHIFT planner framework, which integrates semantic mapping, adaptive coverage planning, and real-time obstacle avoidance to enable comprehensive coverage across diverse terrains and semantic attributes. 
We first develop an innovative Radiant-Field-Informed Coverage Planning (RFICP) algorithm, which generates trajectories that adapt to terrain variations by aligning with environmental changes. Additionally, a Gaussian diffusion field is employed to adaptively adjust the robot's speed, ensuring efficient and uniform coverage under varying environmental conditions influenced by target semantic attributes.
Next, we present a novel incremental KD-tree sliding window optimization (IKD-SWOpt) method to effectively handle dynamic obstacles. IKD-SWOpt leverages an enhanced A* algorithm guided by the IKD-tree distance field to generate initial local avoidance trajectories. Subsequently, it optimizes trajectory segments within and outside waypoint safety zones by evaluating and refining non-compliant segments using an adaptive sliding window. This method not only reduces computational overhead but also guarantees high-quality real-time obstacle avoidance.
Extensive experiments in both simulated and real environments are conducted using drones and robotic vacuum cleaners. The SHIFT planner demonstrates state-of-the-art performance in coverage uniformity and adaptability across various terrains while maintaining very low computational overhead.
\end{abstract}

\section{Introduction}

The increasing reliance on autonomous robots across various fields has highlighted the need for advanced algorithmic frameworks that ensure both efficiency and comprehensive coverage of the task environment~\cite{megalingam2025cleaning, hafeez2023implementation, mier2023indoor}. However, most existing methodologies prioritize minimizing traversal time and path length, often overlooking the critical need for uniform coverage and adaptability to dynamic conditions. Applications such as surface wiping~\cite{megalingam2025cleaning}, uniform pesticide spraying~\cite{hafeez2023implementation}, and autonomous room cleaning~\cite{kang2014robust} require optimized path coverage to achieve efficient and thorough results. Despite their widespread use, existing algorithms fail to fully address the task-specific coverage requirements—whether ensuring that the entire surface is cleaned thoroughly or maintaining accurate 3D reconstruction for inspections~\cite{feng2024fc}. More comprehensive task models are essential for improving the real-world performance of these systems.

Coverage path planning (CPP) is a fundamental problem in robotics, aiming to determine a path for a robot to cover an entire space~\cite{galceran2013survey}.
Classical methods, such as Boustrophedon Cellular Decomposition (BCD)~\cite{choset1998coverage} and Rapidly-Exploring Random Trees (RRT)~\cite{lavalle1998rapidly}, only ensure basic coverage feasibility and focus purely on geometric or kinematic aspects in 2D or pseudo-2D spaces.
Over time, more sophisticated optimization-based CPP algorithms have emerged to balance coverage quality with runtime efficiency and to handle large, unstructured and dynamic environments with real-time constraints and challenging coverage objectives~\cite{cao2020hierarchical, brown2023cdm, marine2023spiral, feng2024fc}. 

However, most existing coverage algorithms rely on assumptions of simplified environments, often neglecting environmental attributes and 3D morphological factors.
For instance, deformable spiral coverage algorithms in~\cite{marine2023spiral} generate smooth coverage paths through Deformable Spiral Coverage Path Planning (DSCPP), but they do not fully account for 3D morphological surfaces or cost-based cleaning validation. 
Similarly, the trochoid-based path planning algorithm for fixed-wing UAVs under wind disturbances proposed in~\cite{fixedwing2023wind} focuses solely on optimizing flight time, without considering terrain or mission-level objectives. 
Most existing offline coverage planning algorithms, such as Boustrophedon, grid-based TSP, and contour-line-based coverage, fail to incorporate real-time adjustments and semantic-level costs~\cite{mier2023indoor}.
The primitive-based path planning approach, which utilizes path primitive sampling and coverage graphs to optimize distance and time, also does not fully address environmental attributes and 3D morphological factors~\cite{cpp2023visual}. 
In~\cite{cao2020hierarchical}, a hierarchical decomposition approach is developed to divide the environment into manageable subregions. However, this method lacks explicit semantic or dynamic updates, resulting in feasible but suboptimal solutions in challenging circumstances. 

Moreover, complex dynamic environments necessitate modern coverage strategies that can react to changes or obstacles in real-time through local re-planning and time allocation. In response to this requirement, \cite{glasius2023online} developed bio-inspired neural-network-based genetic algorithms capable of adapting coverage on-the-fly. However, these algorithms lack considerations for path continuity and time optimality \cite{cao2024learning}.
Similarly, \cite{brown2023cdm} proposed a constriction decomposition method that focuses on time-optimized coverage and 3D reconstruction but does not explicitly model cleaning tasks as a function of coverage cost.
Additionally, the open-source library Fields2Cover~\cite{mier2023fields2cover} provides a comprehensive CPP framework with ready-to-use coverage libraries supporting various types of curvature turns (e.g., Dubins and Reeds-Shepp). Nevertheless, it remains limited to 2D terrain and cannot address real-time obstacle avoidance or dynamic re-planning of smooth paths.

In conclusion, while many notable efforts have advanced coverage in specialized domains, the integration of 3D terrain, real-time re-planning, comprehensive semantic modeling (e.g., dirtiness as a cost function), and dynamic obstacle avoidance remains an active research challenge. Our work addresses these issues by proposing a unified framework, the SHIFT planner, which leverages a field-informed coverage approach with dynamic attribute-based speed control and an incremental sliding window optimization for real-time obstacle avoidance. Unlike the aforementioned methods that simplify 3D structures or treat complete coverage as a purely geometric challenge, our approach utilizes both local and global environmental attributes to regulate coverage speed and thoroughly consider obstacles and geometric constraints. 
Through the proposed RFICP and IKD-SWOpt approaches, we aim to ensure consistent coverage quality across non-uniform 3D terrain, maintain adaptability and path continuity in dynamic environments, and leverage semantic-level cues for cost-based planning. Extensive simulations and real-world experiments demonstrate the state-of-the-art performance of our SHIFT planner in terms of coverage uniformity and adaptability, while maintaining low computational overhead.
 
This paper is organized as follows. Section~\ref{sec:system_overview} presents the overall system architecture of the SHIFT planner. In Section~\ref{sec:surface_extraction}, the procedure for smooth surface modeling and representation is introduced.
Sections~\ref{sec:direct planning}, \ref{sec:RFICP} and \ref{sec:ikdswopt} present the details of our methodology and speed planning strategies. Then, the implementation details of SHIFT planner is provided in Section~\ref{sec:shiftplanner_overall}. Experimental results and analyses are given in Section~\ref{sec:results}. Conclusions and future work are presented in Section~\ref{sec:conclusion}.

\section{System Overview}
\label{sec:system_overview}

\begin{figure*}[ht!]
\centering
\includegraphics[width=\textwidth]{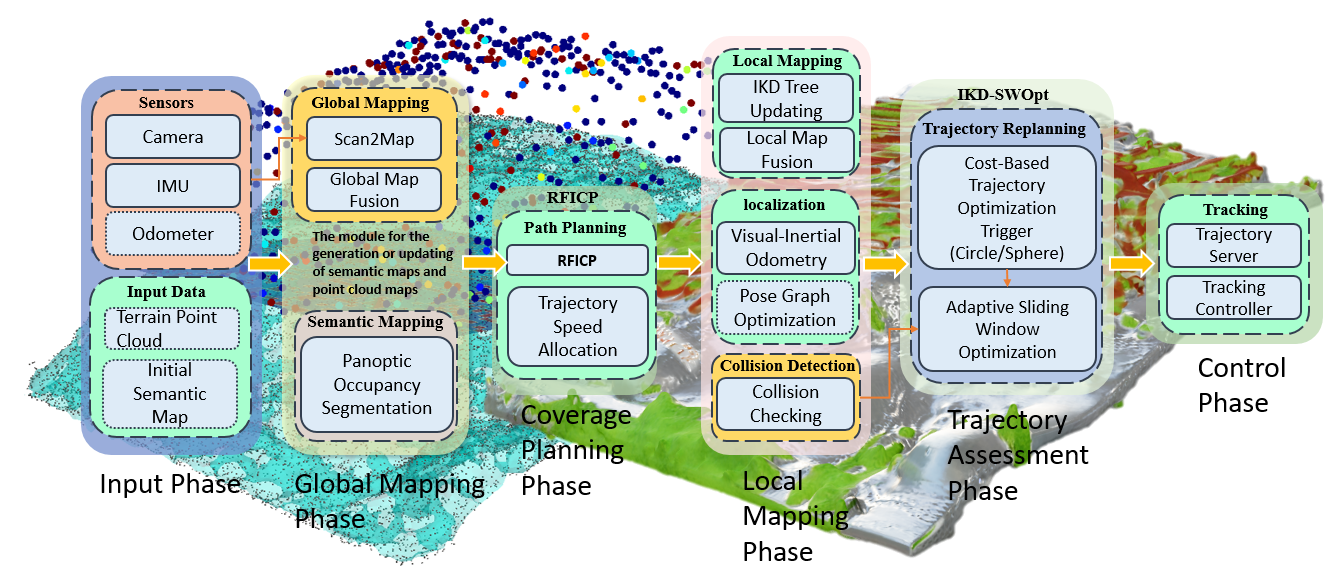}
\caption{Overall architecture of the SHIFT planner. Camera and LiDAR data are combined with an AI-based semantic map and robust VINS pose estimates to form a semantically enriched surface space. RFICP then generates an initial terrain-adaptive coverage trajectory, while IKD-SWOpt refines local paths in real time when encountering obstacles.}
\label{fig:system_architecture}
\end{figure*}

The proposed SHIFT planner framework aims to achieve semantically aware coverage and real-time low-overhead obstacle avoidance in 3D environments. As illustrated in Fig.~\ref{fig:system_architecture}, SHIFT operates in two key stages: RFICP and IKD-SWOpt.
First, the terrain point cloud is fused with local semantic information. An AI-based recognition module provides attributes such as dirtiness or dryness, while a visual-inertial system (VINS) offers robust pose estimates. These data streams jointly form a \emph{surface space} encoding both the geometric and semantic properties of the environment, as described in Section~\ref{sec:RFICP}. Based on these fused inputs, the RFICP module generates a coverage trajectory that smoothly adapts to 3D surface variations. Additionally, using a Gaussian diffusion field, RFICP regulates the robot’s speed according to semantic importance, such as spending more time on highly soiled areas.
Next, to handle dynamic obstacles, IKD-SWOpt is applied for local trajectory refinement, as illustrated in Section~\ref{sec:ikdswopt}. This process begins with an A* search guided by an incremental KD-tree distance field, yielding a promising local avoidance path. IKD-SWOpt then scrutinizes segments near each waypoint safety zone. If a segment fails the smoothness check, it is processed through an adaptive sliding window for further optimization. This strategy significantly reduces computational overhead while ensuring rapid and collision-free path updates.
Finally, the resulting trajectories are fed into a tracking controller to ensure accurate execution of the motion.

\section{Surface Extraction for Elevation Mapping}
\label{sec:surface_extraction}

Accurate terrain modeling is essential for elevation-sensitive coverage trajectory generation. We use a differential geometry approach to fit a smooth surface to the sensor-derived point cloud, then compute and filter curvature metrics (Gaussian and mean curvatures) to remove noise, slender obstacles, or protrusions that do not contribute to meaningful coverage. This yields a continuous surface representation that can be efficiently queried for elevation values during path planning.

Given a set of 3D points sampled from LiDAR or depth sensors, a parametric surface can be fitted as:
\begin{equation}
    \mathbf{S}(u, v) \;=\; 
    \begin{bmatrix}
        x(u, v) \\
        y(u, v) \\
        z(u, v)
    \end{bmatrix},
\label{eq:param_surface}
\end{equation}
where $(u, v)$ parameterize the surface domain via spline fitting or local patch approximation. The first partial derivatives with respect to these parameters can be denoted as $\mathbf{S}_u$ and $\mathbf{S}_v$, and the second partial derivatives are represented as $\mathbf{S}_{uu}, \mathbf{S}_{uv}, \mathbf{S}_{vv}$.

The first and second fundamental forms are used to rigorously evaluate surface shape. The first fundamental form encodes local surface stretching or tangential metric:
\begin{equation}
I \;=\; E\,du^2 \;+\; 2F\,du\,dv \;+\; G\,dv^2,
\end{equation}
where the coefficients can be computed as:
\begin{equation}
E \;=\; \mathbf{S}_u \cdot \mathbf{S}_u,\quad
F \;=\; \mathbf{S}_u \cdot \mathbf{S}_v,\quad
G \;=\; \mathbf{S}_v \cdot \mathbf{S}_v.
\end{equation}
The second fundamental form relates to local curvature behavior in the normal direction:
\begin{equation}
II \;=\; L\,du^2 \;+\; 2M\,du\,dv \;+\; N\,dv^2,
\end{equation}
where the coefficients can be computed as
\begin{equation}
L \;=\; \mathbf{S}_{uu} \cdot \mathbf{n},\quad
M \;=\; \mathbf{S}_{uv} \cdot \mathbf{n},\quad
N \;=\; \mathbf{S}_{vv} \cdot \mathbf{n},
\end{equation}
and the unit normal vector $\mathbf{n}$ is given by
\begin{equation}
\mathbf{n} \;=\; \frac{\mathbf{S}_u \times \mathbf{S}_v}{\|\mathbf{S}_u \times \mathbf{S}_v\|}.
\end{equation}
The Gaussian curvature $K$ and the mean curvature $H$, which measure the intrinsic bending and overall shape of the surface patch respectively, can be obtained as
\begin{equation}
K \;=\; \frac{\,L\,N \;-\; M^2\,}{\,E\,G \;-\; F^2\,},\hspace{1mm}
H \;=\; \frac{\,E\,N \;+\; G\,L \;-\; 2F\,M\,}{\,2\,(E\,G \;-\; F^2)\,},
\label{eq:curvatures}
\end{equation}
Large or abrupt curvature values often indicate noise, protrusions, or slender obstacles on the terrain.

Although the above representation captures the terrain shape, raw sensor data may contain outliers or small-scale obstacles irrelevant to coverage. To detect and filter such artifacts, thresholds of the Gaussian curvature $K_{\text{th}}$ and the mean curvature $H_{\text{th}}$ are defined as
\begin{align}
    K_{\text{th}} &= \mu_{K} + \alpha\,\sigma_{K}, \quad
    H_{\text{th}} = \mu_{H} + \beta\,\sigma_{H},
\end{align}
where $\alpha$ and $\beta$ are tunable hyper parameters to balance noise removal and feature preservation, and the mean $\mu$ and standard deviation $\sigma$ of $K$ and $H$ are calculated over the entire surface.
Any point where $|K|>K_{\text{th}}$ or $|H|>H_{\text{th}}$ is identified as a candidate outlier.
We can then apply either Laplacian smoothing or a removal heuristic to the candidate point $\mathbf{S}(u,v)$ in the form,
\begin{equation}
    \mathbf{S}_{\text{new}}(u,v) 
    \;=\; \mathbf{S}(u,v) \;+\; \lambda\,\Delta \mathbf{S}(u,v),
\end{equation}
where $\Delta \mathbf{S}$ is the Laplacian operator approximated by the difference between the point and the average of its neighbors, and $\lambda$ is a small factor controlling smoothing strength. The aforementioned outlier detection and smoothing processes are performed iteratively until curvature statistics converge below thresholds $K_{\text{th}}$ and $H_{\text{th}}$.

After obtaining a sufficiently smooth surface, we discretize the parametric domain to extract elevation values. Then each discretized grid point $(u_i, v_j)$ on the surface is evaluated in Eqn. \eqref{eq:param_surface} to obtain the position $(x_i,y_j,z_i)$, which is stored as a 2D elevation map $z_i(x_i, y_i)$ for downstream coverage planning. Optionally, bilinear or bicubic interpolation is applied to refine spatial resolution. The resulting elevation map enables path planners to query the height at arbitrary $(x,y)$ during trajectory generation.


\section{Direct Landmark Point Planning}
\label{sec:direct planning}

To achieve efficient and complete coverage of the terrain, we propose a direct landmark point planning method that discretizes the smoothed parametric surface into a uniform grid of points in the parameter domain. This method avoids traditional cell decomposition and is easier to implement, naturally adapting to the terrain's geometry.

We first discretize the parameter domain of the smoothed surface $\mathbf{S}(u, v)$ to create a uniform grid of parameters $\{(u_i, v_j)\}$, which forms the basis for placing landmark points on the terrain. Assuming $u \in [u_{\text{min}}, u_{\text{max}}]$ and $v \in [v_{\text{min}}, v_{\text{max}}]$, we define:

\begin{equation}
u_i = u_{\text{min}} + i \Delta u, \quad v_j = v_{\text{min}} + j \Delta v ,
\end{equation}
where $i = 0, 1, \ldots, N_u$, $j = 0, 1, \ldots, N_v$, and $\Delta u$, $\Delta v$ are the parameter increments.
Each grid point $(u_i, v_j)$ is then mapped onto the 3D surface to obtain the corresponding landmark point,

\begin{equation}
\mathbf{p}_{i,j} = \mathbf{S}(u_i, v_j) = \begin{bmatrix}
x(u_i, v_j) \\
y(u_i, v_j) \\
z(u_i, v_j)
\end{bmatrix}.
\end{equation}

In many coverage tasks like cleaning and spraying, we assume that the coverage action occurs at a constant vertical distance above the local terrain. This ensures that the effect on the underlying semantic attributes remains consistent. Hence, we add a fixed offset $z_{\text{offset}}$ to the actual terrain height $z(x,y)$:

\begin{equation}
\mathbf{p}_{i,j}^{\text{adj}}
\;=\; 
\mathbf{p}_{i,j} \;+\;
\begin{bmatrix}
    0\;0\;z_{\text{offset}}
\end{bmatrix}^T,
\end{equation}
where $\mathbf{p}_{i,j}$ denotes the original waypoint on the extracted 3D surface, and $\mathbf{p}_{i,j}^{\text{adj}}$ is the shifted waypoint. By maintaining this offset, the planner generates coverage trajectories that run parallel to the local topography while preserving consistent coverage impact over all regions.
Finally, we arrange the adjusted landmark points $\mathbf{p}_{i,j}^{\text{adj}}$ in a boustrophedon (zigzag) pattern to ensure complete and systematic coverage of the terrain.


\section{Radiant-Field-Informed Coverage Planning}
\label{sec:RFICP}

After extracting a smooth terrain surface and generating an initial sequence of coverage waypoints as illustrated in the previous section, we incorporate semantic attributes to guide both the coverage path layout and velocity allocation.
Given the terrain-following waypoints $\{\mathbf{p}_{i,j}^{\text{adj}}\}$, the desired coverage effort is defined as
\begin{equation}
    \label{eq:desired coverage}
    C_{\text{desired}}(\mathbf{p}) = k \cdot A(\mathbf{p}),
\end{equation}
where $k$ is the proportional coefficient, and $A(\mathbf{p})$ denotes the semantic field. Regions with a higher $A(\mathbf{p})$ require a proportionally greater coverage effort. Our objective is to ensure that, after passing each waypoint, the environmental attribute is reduced to a target level $C_{\text{target}}$ by effective coverage,
\begin{equation}
    \label{eq:eff_cov_req}
    C_{\text{effective}}(\mathbf{p})
    \;=\;
    C_{\text{desired}}(\mathbf{p}) - C_{\text{target}}
    \;=\;
    k \cdot A(\mathbf{p}) - C_{\text{target}}.
\end{equation}

Then, we assume a Gaussian diffusion kernel to represent how coverage at the robot position $\mathbf{p}'$ influences the environment around position $\mathbf{p}$:
\begin{equation}
    \label{eq:gaussian_kern_iros}
    G(\mathbf{p}; \mathbf{p}') \;=\;
    \frac{1}{\sqrt{2\pi} \sigma} \exp\left(-\frac{\|\mathbf{p} - \mathbf{p}'\|^2}{2\sigma^2}\right),
\end{equation}
where $\|\mathbf{p}-\mathbf{p}'\|\le R$. For computational efficiency, we only consider this influence within a circle of radius $R$. When the robot dwells for time $t$ at position $\mathbf{p}'$, its total coverage at position $\mathbf{p}$ can be modeled as an integration of the Gaussian kernel within the circle:
\begin{equation}
\label{eq:coverage integration}
    C_{\text{diff}}(\mathbf{p}) 
    =
    \int_{\|\mathbf{p}-\mathbf{p}'\|\le R}
    G(\mathbf{p}; \mathbf{p}')\,\left[\,1 - e^{-\lambda\,t(\mathbf{p}')} \right]
    d\mathbf{p}',
\end{equation}
where $\lambda$ is a coverage-efficiency constant. We require $C_{\text{diff}}(\mathbf{p}) = C_{\text{effective}}(\mathbf{p})$ to reduce the environmental attribute to the target level.

Assuming an approximately uniform dwell time $t(\mathbf{p})$ within the local Gaussian footprint, and by combining \eqref{eq:eff_cov_req} and \eqref{eq:coverage integration}, we obtain:
\begin{equation}
\label{eq:almost_final_iros}
\begin{split}
    k\,A(\mathbf{p}) - C_{\text{target}}&=
    \left[\int_{\|\mathbf{p}-\mathbf{p}'\|\le R} G(\mathbf{p}; \mathbf{p}')\,d\mathbf{p}'\right]
    \left[1 - e^{-\lambda\,t(\mathbf{p})}\right]\\
    &=\left[2f\left(\frac{R}{\sigma}\right)-1\right]
    \left[1 - e^{-\lambda\,t(\mathbf{p})}\right].
\end{split}
\end{equation}
where we use the cumulative distribution function (CDF) $f(\frac{R}{\sigma})$ of the standard Gaussian distribution to represent the Gaussian integral for simplicity~\cite{pishro2014introduction}. The dwell time $t(\mathbf{p})$ can then be obtained as
\begin{align}
t(\mathbf{p}) = -\frac{1}{\lambda}\ln\left[
1 - \frac{k \, A(\mathbf{p}) - C_{\text{target}}}{2f\left(\frac{R}{\sigma}\right) - 1}
\right].
\label{eq:tp_final_iros}
\end{align}
This equation is valid only when $k\,A(\mathbf{p}) > C_{\text{target}}$ and the logarithm’s argument is positive. Since inversely proportional to dwell time, the speed $v$ can be calculated as
\begin{align}
v(\mathbf{p}) = \frac{\Delta s}{t(\mathbf{p})}
\approx \frac{1}{t(\mathbf{p})}
= -\frac{\lambda}{
\ln\left[1 - \frac{k\,A(\mathbf{p}) - C_{\text{target}}}
{2f\left(\frac{R}{\sigma}\right) - 1}\right]
}.
\end{align}
where each discretized path segment $\Delta s$ is approximated as unit length. 
We then clip $v(\mathbf{p})$ to lie within velocity limits and enforce acceleration constraints via standard velocity interpolation techniques. 

\section{Real-Time Obstacle Avoidance}
\label{sec:ikdswopt}

While RFICP ensures comprehensive coverage and speed allocation, unforeseen dynamic obstacles can invalidate parts of the trajectory. In our SHIFT planner framework, we achieve real-time obstacle avoidance using the IKD-SWOpt method, which refines local segments upon detection of collision or near-collision states.

\subsection{A* Search with IKD-Tree-Informed Distance Field}

In dynamic or partially known scenarios, a continuously updated incremental k-dimensional tree (IKD-tree) is maintained to represent obstacle locations\cite{cai2021ikd}. We build a distance field $\mathcal{D}(\mathbf{p})$ from the IKD-tree, which represents the distance to the nearest obstacle at point $\mathbf{p}$. Given the distance field $\mathcal{D}$, the initial A* search uses the cost function below:
\begin{equation}
\label{eq:A star cost func}
    f(\mathbf{p})
    \;=\;
    g(\mathbf{p}) + h(\mathbf{p}) + \alpha \,\cdot\, \frac{1}{\,\mathcal{D}(\mathbf{p}) + \varepsilon},
\end{equation}
where $g(\mathbf{p})$ and $h(\mathbf{p})$ are the usual path cost and heuristic terms respectively, $\alpha$ is a weighing factor, and $\varepsilon$ is a small constant. The third term in \eqref{eq:A star cost func} penalizes positions close to obstacles. The IKD-tree provides near-lossless distance-field lookups, enhancing A* heuristics and avoiding local minima. Therefore, this enhanced IKD-tree-informed A* algorithm generates a nice initial avoidance path with better clearance. We then use this path as a starting point for subsequent local gradient-based optimization.


\subsection{Non-Compliant Segment Identification}

Let $\mathcal{P}=\{\mathbf{p}_k\}_{k=1}^{N}$ be the initial trajectory obtained by the enhanced A* search. For each waypoint $\mathbf{p}_k$, we define a circular safety region 
\begin{equation}
    \mathcal{C}_k 
    \;=\;
    \Bigl\{\mathbf{p} \,\Big|\;\|\mathbf{p}-\mathbf{p}_k\|\le r_{\mathrm{safe}}\Bigr\},
\end{equation}
where the radius $r_{\mathrm{safe}}$ can be obtained as the distance to the nearest obstacle in the IKD-tree.

\begin{figure}[t]
    \centering
    \includegraphics[width=0.9\columnwidth]{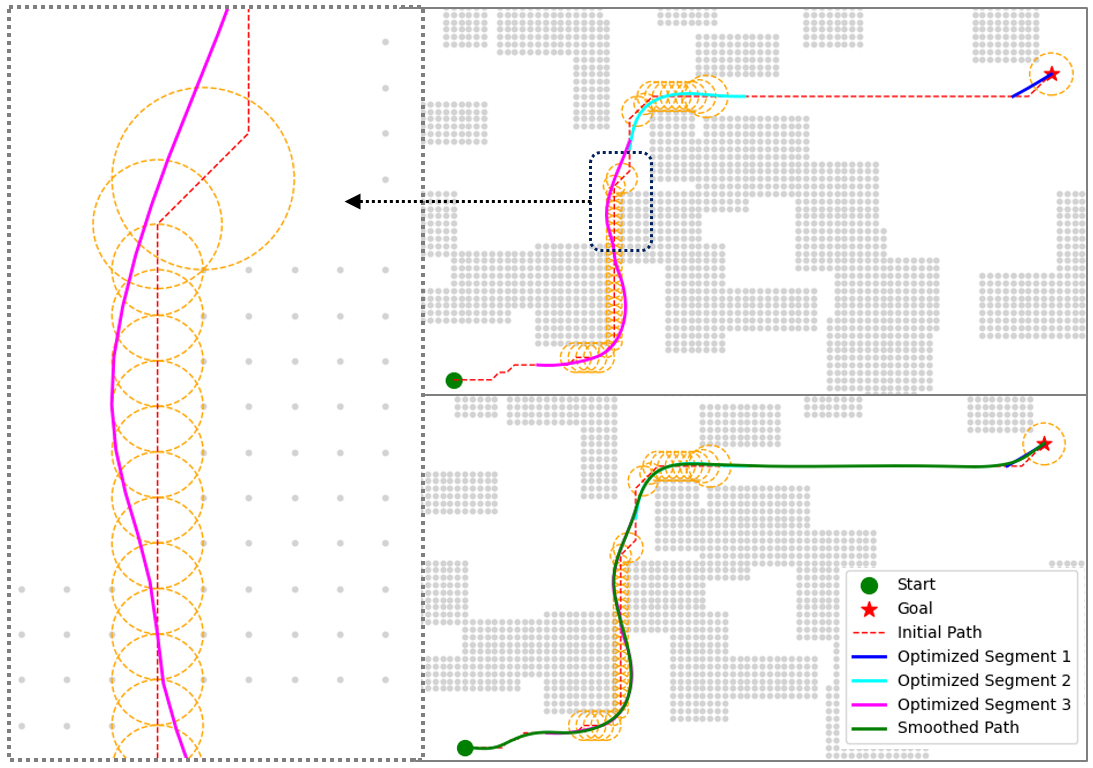}
    \caption{An illustration of the local segment identification and refinement. For each waypoint on the initial path (red), a circular region (yellow) is defined. We then gather all identified non-compliment segments whose safety score exceeds the threshold into an adaptive sliding window for local optimization.}
    \label{fig:circular_judge}
\end{figure}

Given all nearby waypoints in the circular safety region, we define the safety score for $\mathbf{p}_k$ as
\begin{equation}
\label{eq:safety_score}
S_k = \sum\limits_i f[\mathcal{D}(\mathbf{p}_i),C_{\mathrm{cont}}(\mathbf{p}_i),C_{\mathrm{feas}}(\mathbf{p}_i)],\;\text{for}\;\mathbf{p}_i \in \mathcal{C}_k ,
\end{equation}
where $f(\cdot)$ is a weighted combination function, $\mathcal{D}$ represents the distance to the nearest obstacle, $C_{\mathrm{cont}}$ denotes the difference in headings or curvatures between consecutive waypoints, and $C_{\mathrm{feas}}$ is a feasibility metric indicating collisions or violations of dynamic constraints, such as velocity or acceleration limits.
If $S_k$ falls \emph{below} a specified safety threshold $\tau_{\mathrm{safe}}$, the local path segment is considered \emph{non-compliant}. We then mark the surrounding trajectory segment within the circular safety region, denoted by $\mathcal{P}_{k-\Delta:k+\Delta}$, for local re-optimization. As shown in Fig.\ref{fig:circular_judge}, this method ensures that both the immediate vicinity of $\mathbf{p}_k$ and its neighboring waypoints are jointly refined. All identified non-compliant segments are gathered into an \emph{adaptive sliding window} for subsequent optimization. The outline of the non-compliant segment identification is presented in Algorithm~\ref{alg:S_nc identification}.

\begin{algorithm}[ht!]
\small
\caption{Non-Compliant Segment Identification}
\label{alg:S_nc identification}
\begin{algorithmic}[1]
\STATE $\mathcal{S}_{\mathrm{nc}} \leftarrow \emptyset$ 
\FOR{each $\mathbf{p}_k \in \mathcal{P}$}
    \STATE Define $\mathcal{C}_k = \{\mathbf{p}\,\mid\, \|\mathbf{p}-\mathbf{p}_k\|\le r_{\mathrm{safe}}\}$
    \STATE Extract waypoints nearby $\{\mathbf{p}_i \in \mathcal{C}_k\}$
    \STATE Compute safety score $S_k$ via \eqref{eq:safety_score}
    \IF{$S_k < \tau_{\mathrm{safe}}$}
        \STATE $\mathcal{P}_{k-\Delta:k+\Delta} \leftarrow \{\mathbf{p}_i \in \mathcal{C}_k\}$
        \STATE $\mathcal{S}_{\mathrm{nc}} \leftarrow \mathcal{S}_{\mathrm{nc}} \,\cup\, \{\mathcal{P}_{k-\Delta:k+\Delta}\}$
    \ENDIF
\ENDFOR
\RETURN $\mathcal{S}_{\mathrm{nc}}$
\end{algorithmic}
\end{algorithm}


\subsection{IKD-Tree Sliding Window Optimization}

Given that each segment already has a well-informed initial path obtained from the A* search, our solver refines only the control points within the window while maintaining global continuity. The local cost function is given by
\begin{equation}
\label{eq:swopt_cost}
    J(\mathbf{x})=J_{\mathrm{obs}}(\mathbf{x})+
    J_{\mathrm{smooth}}(\mathbf{x})+J_{\mathrm{len}}(\mathbf{x}),
\end{equation}
denoting obstacle penalty, continuity and path efficiency respectively. We then employ a gradient-based solver, such as the limited-memory BFGS (L-BFGS) method~{\cite{liu1989limited}}, to minimize~\eqref{eq:swopt_cost} for the waypoints within the sliding window. This effectively pushes the path away from obstacles while maintaining coverage constraints. Additionally, this segmented optimization is highly parallelizable and can leverage GPU acceleration. By distributing the optimization of segments across GPU cores, we can significantly reduce computation time and improve real-time obstacle avoidance in dynamic environments.

Each time a window is optimized, the updated segment is merged back into the global trajectory. We then apply B-spline fitting to smooth the transition between original and optimized segments, and the smoothed trajectory is given by
\begin{equation}
    \mathbf{C}(t)
    \;=\;
    \sum_{i=0}^{n}
    \mathbf{p}_i \,N_{i,k}(t),
\end{equation}
where $\mathbf{p}_i$ represents the combined control points from both segments, and $N_{i,k}$ is the base function. This continuous spline ensures feasible velocities and accelerations at junctions and avoids abrupt turns. The general flow of the IKD-SWOpt algorithm is presented in Algorithm~\ref{alg:IKDSWOpt}. It efficiently handles dynamic obstacles by incrementally updating the KD-tree with new obstacle information and refining only the affected segments of the trajectory.

\begin{algorithm}[ht!]
\caption{IKD-SWOpt}
\small
\label{alg:IKDSWOpt}
\begin{algorithmic}[1]

\WHILE{new obstacle data arrives}
    \STATE \textbf{Update IKD-tree}: Insert new obstacles into $\mathcal{T}$
    \STATE \textbf{Segment Identification}: Obtain $\mathcal{S}_{\mathrm{nc}}$ from Algorithm~\ref{alg:S_nc identification}
    \FOR{each segment $\mathcal{S} \in \mathcal{S}_{\mathrm{nc}}$}
        \STATE \textbf{Sliding Window Optimization}: Optimize $\mathcal{S}$ by minimizing $J(\mathbf{x})$ in \eqref{eq:swopt_cost}
    \ENDFOR
    \STATE \textbf{Trajectory Reconnection}: Merge optimized segments into initial path $\mathcal{P}_0$ using B-spline smoothing
\ENDWHILE

\RETURN $\mathcal{P}_{\mathrm{opt}}$
\end{algorithmic}
\end{algorithm}

\section{SHIFT Planner Implementation}
\label{sec:shiftplanner_overall}

The outline of the SHIFT planner is shown in Algorithm~\ref{alg:shift_planner}. Terrain-aware coverage planning (RFICP) is combined with incremental obstacle avoidance (IKD-SWOpt). First, a smooth surface is extracted from raw LiDAR data and discretized into a grid of waypoints, offset for uniform coverage height, and then sequenced in a boustrophedon pattern. In RFICP, an initial coverage trajectory is generated with semantic-driven speeds. When obstacles change, an incremental KD-tree updates the distance field. In IKD-SWOpt, non-compliant segments are first initialized using local A* search with IKD-tree heuristics, identified by circular safety region check, and refined in an adaptive sliding window. This framework ensures obstacle clearance, curvature smoothness, and path efficiency. Finally, B-spline interpolation is used to reconnect the segments, ensuring global continuity and uniform coverage.

\begin{algorithm}[ht!]
\small
\caption{SHIFT Planner}
\label{alg:shift_planner}
\begin{algorithmic}[1]

\vspace{2pt}
\STATE \textbf{SurfaceExtraction}(\texttt{point cloud}): 
\STATE \quad (a) Compute curvature and filter outliers
\STATE \quad (b) Smooth parametric surface $\mathbf{S}(u,v)$
\STATE \quad (c) Build an elevation map
\vspace{2pt}

\STATE \textbf{LandmarkPointPlanning}():
\STATE \quad (a) Discretize $(u,v)$ into grid $\{\mathbf{p}_{i,j}\}$
\STATE \quad (b) Apply offset $z_{\text{offset}}$, sequence in Boustrophedon pattern
\vspace{2pt}

\STATE \textbf{RFICP}(\texttt{waypoints}, \texttt{semantic map}):
\STATE \quad (a) Evaluate $A(\mathbf{p})$ for each waypoint
\STATE \quad (b) Compute dwell times via Gaussian diffusion
\STATE \quad (c) Allocate velocities $v(\mathbf{p})$, build initial path $\mathcal{P}_0$
\vspace{2pt}

\STATE \textbf{InitIKDtree}():
\STATE \quad (a) Insert known obstacles into IKD-tree $\mathcal{T}$
\STATE \quad (b) Maintain distance field $\mathcal{D}(\mathbf{p})$
\vspace{2pt}

\STATE \textbf{IKD-SWOpt}(\texttt{path} $\mathcal{P}_0$, \texttt{ikdtree} $\mathcal{T}$):
\STATE \quad \textbf{while} \textit{obstacle updates} exist \textbf{do}
\STATE \quad \quad (a) Local A* with IKD-tree-informed heuristic
\STATE \quad \quad (b) Identify $\mathcal{S}_{\mathrm{nc}}$ via circular judgment
\STATE \quad \quad (c) SlidingWindowOpt with BFGS refinement
\STATE \quad \quad (d) B-spline reconnection
\STATE \quad \textbf{end while}
\STATE \quad \textbf{return} $\mathcal{P}_{\mathrm{opt}}$
\vspace{2pt}

\STATE \textbf{TrajectoryOutput}:
\STATE \quad $\mathcal{P}_{\mathrm{final}} \leftarrow \text{IKD-SWOpt}(\mathcal{P}_0,\mathcal{T})$
\STATE \quad \textbf{return} $\mathcal{P}_{\mathrm{final}}$
\end{algorithmic}
\end{algorithm}


\section{Results}
\label{sec:results}
The proposed SHIFT planner has been validated in both simulated and physical experiments. Our experimental setup and implementation details are described in Section \ref{sec:experiment}. Then we compare its coverage performance under various semantic attributes and real-time capability of local trajectory optimization in the presence of obstacles with a few benchmark state-of-the-art planning algorithms in Section \ref{sec:coverage performance} and Section \ref{sec:local planning}, respectively.

\subsection{Experimental Setup}
\label{sec:experiment}
For both numerical and physical experiments, all point cloud processing, including surface extraction and curvature filtering, is implemented in C++ for real-time efficiency. The IKD-tree is updated dynamically as new obstacle data arrives, providing near-lossless distance queries for local re-planning. The experiments are carried out on a system equipped with an Intel i7 CPU at 3.4\,GHz and 16\,GB RAM.

\begin{figure*}[ht]
    \centering
    \begin{subfigure}[b]{0.3\textwidth}
        \centering
        \includegraphics[width=\textwidth]{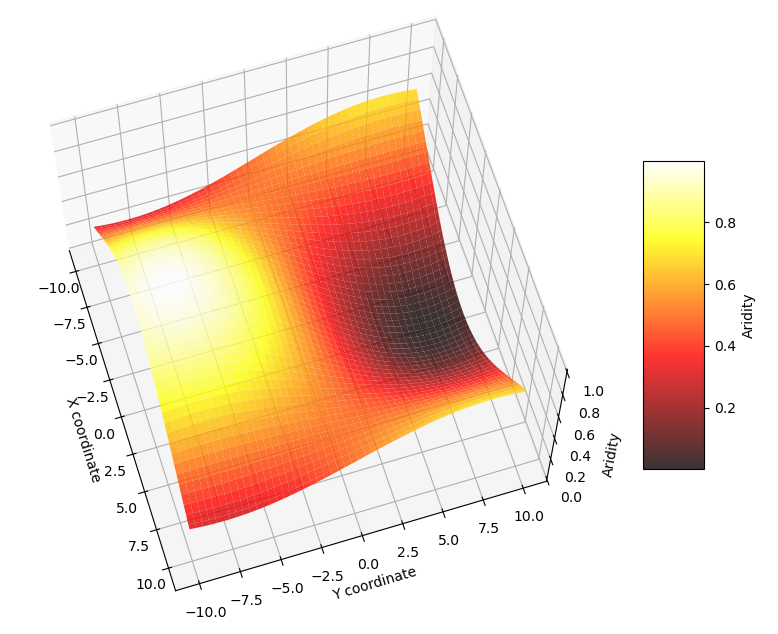}
        \caption{Aridity Semantic Map}
        \label{fig:arid_map}
    \end{subfigure}
    \hspace{2mm}  
    \begin{subfigure}[b]{0.34\textwidth}
        \centering
        \includegraphics[width=\textwidth]{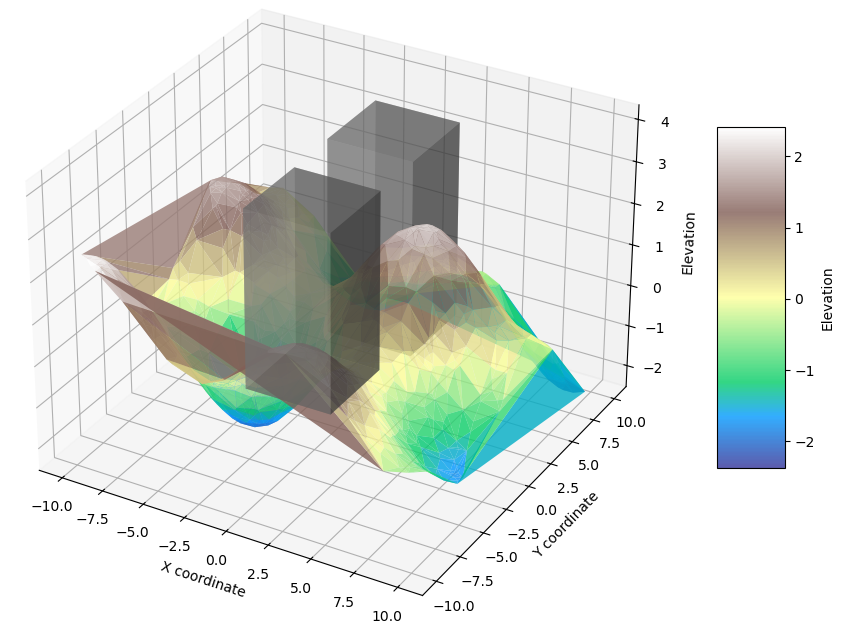}
        \caption{Obstacle Configuration}
        \label{fig:obstacle_map}
    \end{subfigure}
    \hspace{2mm}  
    \begin{subfigure}[b]{0.3\textwidth}
        \centering
        \includegraphics[width=\textwidth]{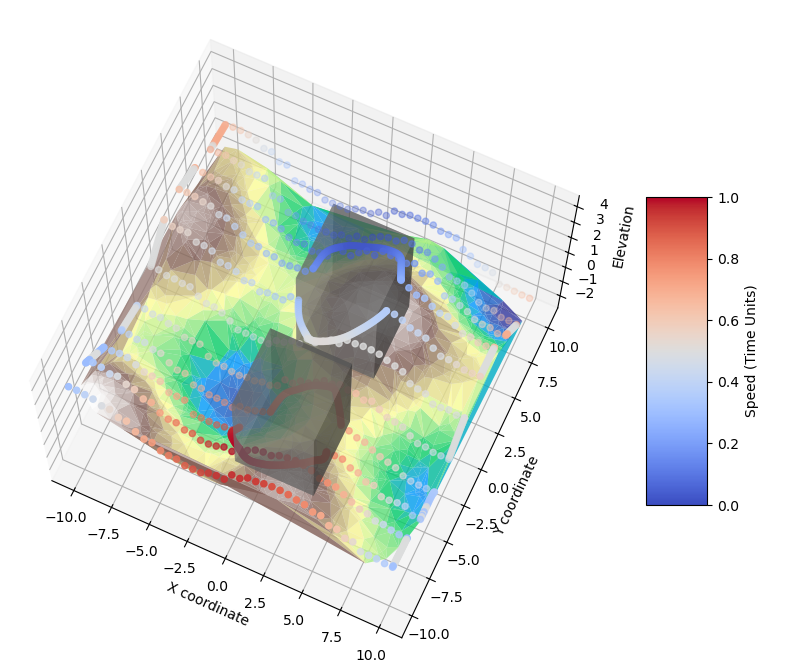}
        \caption{Coverage Trajectory}
        \label{fig:ikd_swopt}
    \end{subfigure}
    \caption{Numerical simulation results: the SHIFT planner generates coverage trajectory that adapts to the environment and refines locally to avoid obstacles.}
    \label{fig:rfibc_trajectory}
\end{figure*}

In numerical experiments, an agricultural drone is simulated to conduct coverage tasks in outdoor terrains with non-uniform dryness levels. Its onboard LiDAR and camera provide terrain point clouds and semantic information via sensor fusion and segmentation. We construct a series of simulated maps, which contain gradient or patch-based variations in semantic attributes (Fig.~\ref{fig:arid_map}), and multiple static barriers, narrow corridors or dynamic moving objects (Fig.~\ref{fig:obstacle_map}). Fig.~\ref{fig:ikd_swopt} demonstrates how IKD-SWOpt method locally refines the path to bypass obstacles and adjusts its speed according to local aridity level in real time to uniformly irrigate drier crops.

\begin{figure}[ht!]
\centering
\includegraphics[width=0.6\linewidth]{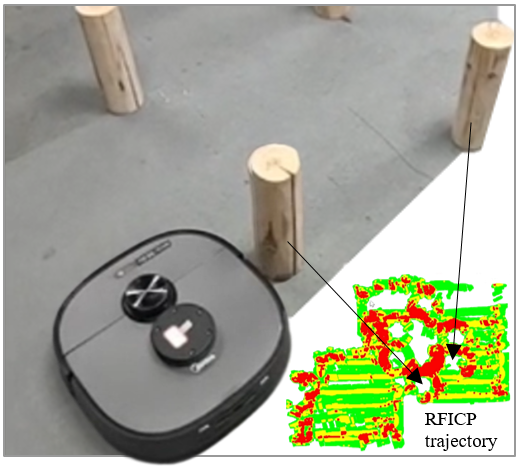}
\caption{A vacuum robot navigates a semantic-labeled indoor environment, slowing down in dirtier zones (red).}
\label{fig:real_robot_semantic_dirtiness}
\end{figure}

Our SHIFT planning method is also validated physically on a robotic vacuum cleaner as shown in Fig.~\ref{fig:real_robot_semantic_dirtiness}. Equipped with a LiDAR, camera and IMU for mapping and localization, the robot operates in a dirtiness-labeled indoor environment deployed with some cylindrical obstacles. SHIFT Planner automatically allocates longer dwell times in heavily soiled areas, while IKD-SWOpt swiftly reacts to moving furniture and human obstacles. Both numerical and physical experiments demonstrate that the SHIFT planner provides thorough semantic coverage and low-latency obstacle avoidance, making it ideal for complex real-world conditions.

\subsection{Coverage Performance Analysis}
\label{sec:coverage performance}

The coverage performance of the proposed SHIFT planner is compared with a baseline coverage method with a uniform speed profile, and other state-of-the-art methods including CCPP~\cite{chen2021clustering} and K-Means+DRL~\cite{ni2024cooperative}. This benchmark comparison is done on a robot vacuum cleaner, Eureka J15pro~\cite{EurekaIntro}, navigating in a standard multi-room coverage test environment with heavily soiled areas and common household obstacles such as desks, chairs, and other furniture. Additionally, a dynamic tracking system was integrated to automatically compute the coverage ratio in real-time.

The coverage performance comparison regarding four performance metrics is presented in Table~\ref{table:coverage_comparison}. The \emph{coverage completeness} is defined as the ratio of the area cleaned above a preset threshold to the total area. 
The \emph{path overlap rate} is represented as the proportion of the path re-traversing cleaned areas. 
The \emph{energy efficiency} is defined as the normalized energy consumption per unit area cleaned. 
The \emph{cleaning uniformity} is derived from the normalized cleaning intensity distribution over the test area. A higher value of cleaning uniformity indicates more uniform cleaning.

Due to semantic-driven speed modulation and local optimization, SHIFT planner achieves significantly higher coverage completeness ($13.3\%$ higher than the baseline) and cleaning uniformity ($34.7\%$ higher than the baseline), while maintaining lower path overlap and higher energy efficiency. Although the CCPP and K-Means+DRL methods outperform the baseline, they still fall short of the SHIFT planner in achieving uniform cleaning in dynamic environments. Therefore, SHIFT planner not only ensures better coverage but also reduces redundancy in path planning, making it more efficient overall.

\begin{table}[ht]
\centering
\caption{Coverage Performance Comparison}
\label{table:coverage_comparison}
\resizebox{\linewidth}{!}{%
\begin{tabular}{lcccc}
\hline
\textbf{Metric} & \textbf{SHIFT Planner} & \textbf{Baseline Coverage} & \textbf{CCPP} & \textbf{K-Means+DRL} \\
\hline
Coverage Completeness (\%) & \textbf{98.6} & 85.3 & 96.8 & 94.5 \\
Path Overlap Rate (\%)     & \textbf{4.3}  & 10.2 & 8.7  & 12.1 \\
Energy Efficiency (Norm.)  & \textbf{0.82} & 0.65 & 0.74 & 0.69 \\
Cleaning Uniformity (\%)   & \textbf{95.2} & 60.5 & 90.1 & 88.3 \\
\hline
\end{tabular}
}
\end{table}

\begin{figure}[ht]
\centering
\includegraphics[width=\linewidth]{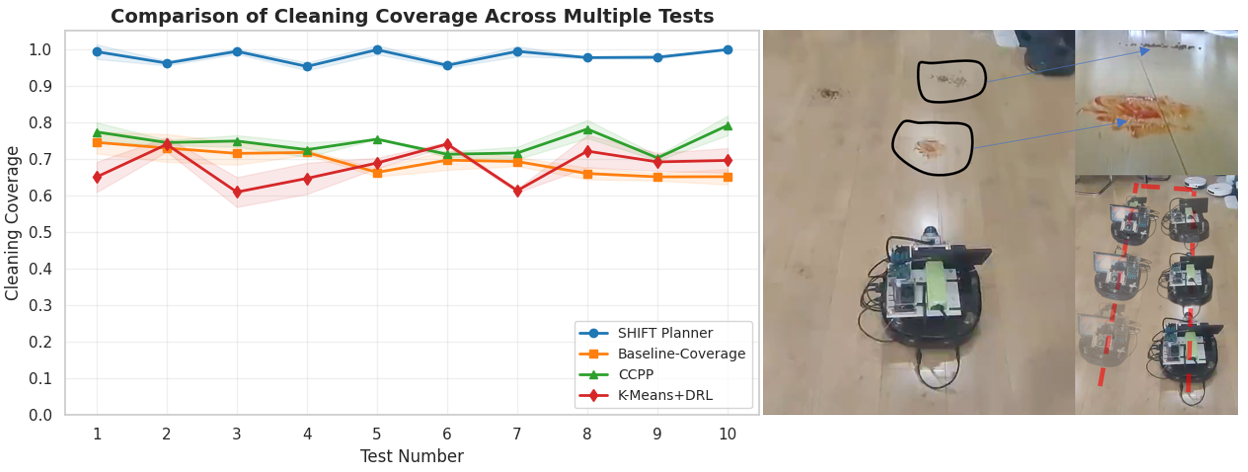}
\caption{Comparison of cleaning uniformity: SHIFT Planner achieves uniform cleaning in high-demand regions, while the other three methods show inadequate cleaning in heavily soiled areas.}
\label{fig:coverage_uniformity}
\end{figure}

To further demonstrate the effectiveness of RFICP, Fig.~\ref{fig:coverage_uniformity} shows the cleaning uniformity across different regions. SHIFT planner ensures that heavily soiled areas circled in black have extended dwell times with slower speed, resulting in thorough cleaning. However, the other three methods fail to adequately clean these regions despite covering the entire area. Among $10$ different cleaning tests, their cleaning coverage rate is approximately $30\%$ lower than the SHIFT planner on average.

\subsection{Local Trajectory Planning Ability}
\label{sec:local planning}

We compare the IKD-SWOpt local trajectory refinement capability with two well-known methods based on Euclidean Signed Distance Field (ESDF), the Fast-Planner~\cite{zhou2019robust} and EWOK~\cite{usenko2017real}. Fast-Planner is an efficient kinodynamic planner with a front-end graph search and a back-end gradient-based optimization over an ESDF. EWOK is a B-spline–based local planner that updates a 3D occupancy buffer and ESDF for MAVs in real time. The comparison is performed on a 3D map with randomly placed cylindrical obstacles of varying heights. With the obstacle density ranging from $0.1$ to $0.5$ obs/m\textsuperscript{2}, the local re-planning capabilities of these three planners are compared in terms of total flight time $t$, path length $L$, energy consumption $E$, and per-iteration planning time $t_{\mathrm{p}}$. Additionally, the performance of IKD-SWOpt with and without GPU acceleration are both evaluated to demonstrate its computational efficiency and scalability. Table~\ref{tab:ikdcomparison} presents the average performance metrics among several runs at a medium obstacle density of $0.3$ obs/m\textsuperscript{2}.

\begin{table}[ht]
\centering
\caption{Comparison of Local Planning Abilities}
\label{tab:ikdcomparison}
\resizebox{\linewidth}{!}{%
\begin{tabular}{lccccc}
\hline
\textbf{Planner} & $t$(s) & $L$(m) & $E$ (Norm.) & $t_{\mathrm{p}}$(ms) & \textbf{GPU Accel.} \\
\hline
EWOK & 22.3 & 38.7 & 189.5 & 3.1 & Off \\
Fast-Planner & 21.6 & 35.4 & 142.2 & 4.2 & Off \\
\textbf{IKD-SWOpt} & 21.9 & 36.2 & 153.8 & 1.6 & Off \\
\textbf{IKD-SWOpt} & 21.5 & 35.9 & 150.3 & 0.9 & On \\
\hline
\end{tabular}
}
\end{table}

IKD-SWOpt without GPU acceleration requires no explicit ESDF construction, reducing the per-iteration planning time $t_{\mathrm{p}}$ to $1.6$ ms, which is less than half of the EWOK and Fast-Planner's planning time on average. With GPU acceleration, $t_{\mathrm{p}}$ is significantly further reduced to $0.9$ ms. The path length $L$ and total flight time $t$ of IKD-SWOpt are comparable to Fast-Planner’s kinodynamic search, indicating efficient path planning without the overhead of repeated ESDF updates. Although IKD-SWOpt without GPU acceleration has $8\%$ higher energy consumption compared to Fast-Planner, the GPU-accelerated version reduces energy cost to $150.3$ normalized units, keeping it competitive. The multi-segment sliding window optimization approach in IKD-SWOpt is highly parallelizable, allowing simultaneous optimization of multiple trajectory segments on GPU cores. This parallelization significantly reduces computation time, enables real-time trajectory adjustments and enhances the planner's responsiveness in dynamic environments.

\begin{figure}[ht]
    \centering
    \includegraphics[width=0.8\linewidth]{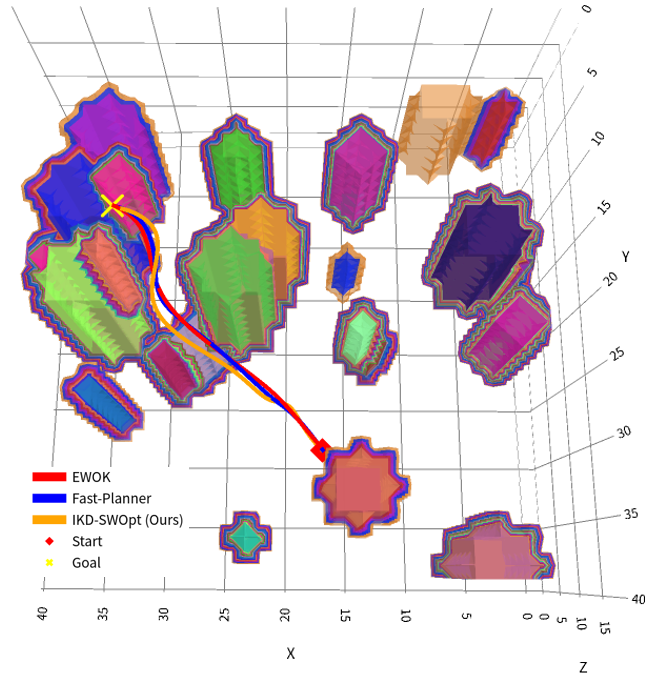}
    \caption{Trajectories of EWOK (red), Fast-Planner (blue), and IKD-SWOpt (orange) at an obstacle density of $0.3$ obs/m\textsuperscript{2}. Our approach avoids excessive detours and eliminates large ESDF updates.}
    \label{fig:comp_trajs}
\end{figure}

Fig.~\ref{fig:comp_trajs} shows the trajectories of IKD-SWOpt, EWOK and Fast-Planner. While Fast-Planner and EWOK generate feasible paths, they frequently trigger ESDF rebuilds, especially in narrow passages. In contrast, IKD-SWOpt uses IKD-tree queries and sliding-window refinement to maintain real-time responsiveness with fewer detours. The GPU-accelerated version further improves efficiency, allowing quick trajectory adjustments even in highly dynamic environments.

\section{Conclusion}
\label{sec:conclusion}

In this paper, we introduce the SHIFT planner, a novel coverage planning framework designed for autonomous robots in complex and dynamic environments. It combines semantic mapping, terrain-aware coverage planning, and dynamic obstacle avoidance to achieve efficient and uniform coverage.

Our RFICP algorithm adjusts speed and trajectory using a diffusion field model, allowing adaptive and consistent coverage across various environmental conditions. Additionally, the IKD-SWOpt efficiently handles obstacle avoidance by locally refining trajectories within a sliding detection window, ensuring safe and responsive navigation.

Extensive experiments demonstrate the efficiency and adaptability of our framework in achieving high-quality coverage under real-world constraints. Comparative analyses indicate that the SHIFT planner outperforms both baseline and state-of-the-art methods in terms of coverage completeness, path efficiency, and energy consumption. In the future, we plan to extend this approach to multi-robot systems, enabling collaborative coverage in larger and more dynamic environments.

\bibliographystyle{ieeetr}
\bibliography{references}

\begin{thebibliography}{10}

\bibitem{megalingam2025cleaning}
R.~K. Megalingam, S.~R.~R. Vadivel, S.~S. Kotaprolu, B.~Nithul, D.~V. Kumar, and G.~Rudravaram, ``Cleaning robots: A review of sensor technologies and intelligent control strategies for cleaning,'' {\em Journal of Field Robotics}, 2025.

\bibitem{hafeez2023implementation}
A.~Hafeez, M.~A. Husain, S.~Singh, A.~Chauhan, M.~T. Khan, N.~Kumar, A.~Chauhan, and S.~Soni, ``Implementation of drone technology for farm monitoring \& pesticide spraying: A review,'' {\em Information processing in Agriculture}, vol.~10, no.~2, pp.~192--203, 2023.

\bibitem{mier2023indoor}
G.~Mier, J.~{a}o Valente, and S.~de~Bruin, ``Indoor coverage path planning: Survey, implementation, analysis,'' {\em Journal of Field Robotics}, vol.~40, no.~5, pp.~765--789, 2023.

\bibitem{kang2014robust}
M.-C. Kang, K.-S. Kim, D.-K. Noh, J.-W. Han, and S.-J. Ko, ``A robust obstacle detection method for robotic vacuum cleaners,'' {\em IEEE Transactions on Consumer Electronics}, vol.~60, no.~4, pp.~587--595, 2014.

\bibitem{feng2024fc}
C.~Feng, H.~Li, M.~Zhang, X.~Chen, B.~Zhou, and S.~Shen, ``Fc-planner: A skeleton-guided planning framework for fast aerial coverage of complex 3d scenes,'' in {\em 2024 IEEE International Conference on Robotics and Automation (ICRA)}, pp.~8686--8692, IEEE, 2024.

\bibitem{galceran2013survey}
E.~Galceran and M.~Carreras, ``A survey on coverage path planning for robotics,'' {\em Robotics and Autonomous systems}, vol.~61, no.~12, pp.~1258--1276, 2013.

\bibitem{choset1998coverage}
H.~Choset and P.~Pignon, ``Coverage path planning: The boustrophedon cellular decomposition,'' in {\em Field and service robotics}, pp.~203--209, Springer, 1998.

\bibitem{lavalle1998rapidly}
S.~LaValle, ``Rapidly-exploring random trees: A new tool for path planning,'' {\em Research Report 9811}, 1998.

\bibitem{cao2020hierarchical}
C.~Cao, J.~Zhang, M.~Travers, and H.~Choset, ``Hierarchical coverage path planning in complex 3d environments,'' in {\em 2020 IEEE International Conference on Robotics and Automation (ICRA)}, pp.~3206--3212, IEEE, 2020.

\bibitem{brown2023cdm}
S.~Brown and S.~L. Waslander, ``The constriction decomposition method for coverage path planning,'' {\em IEEE Robotics and Automation Letters}, vol.~8, no.~4, pp.~3233--3238, 2023.

\bibitem{marine2023spiral}
M.~Turner and S.~Robinson, ``A deformable spiral-based algorithm to smooth coverage path planning for marine growth removal,'' {\em Ocean Engineering}, vol.~250, p.~112045, 2023.

\bibitem{fixedwing2023wind}
A.~Johnson and R.~Lee, ``Flight testing boustrophedon coverage path planning for fixed wing uavs in wind,'' {\em Aerospace Science and Technology}, vol.~158, pp.~1124--1137, 2023.

\bibitem{cpp2023visual}
R.~Clark and L.~Wright, ``Coverage path planning using path primitive sampling and primitive coverage graph for visual inspection,'' {\em Automation in Construction}, vol.~155, p.~104052, 2023.

\bibitem{glasius2023online}
J.~Doe and J.~Smith, ``Online complete coverage path planning of a reconfigurable robot using glasius bio-inspired neural network and genetic algorithm,'' {\em Robotics and Autonomous Systems}, vol.~155, pp.~1043--1058, 2023.

\bibitem{cao2024learning}
M.~Cao, X.~Xu, Y.~Yang, J.~Li, T.~Jin, P.~Wang, T.-Y. Hung, G.~Lin, and L.~Xie, ``Learning dynamic weight adjustment for spatial-temporal trajectory planning in crowd navigation,'' {\em arXiv preprint arXiv:2412.00555}, 2024.

\bibitem{mier2023fields2cover}
G.~Mier, J.~{a}o Valente, and S.~de~Bruin, ``Fields2cover: An open-source coverage path planning library for unmanned agricultural vehicles,'' {\em IEEE Robotics and Automation Letters}, vol.~8, no.~4, pp.~2167--2172, 2023.

\bibitem{pishro2014introduction}
H.~Pishro-Nik, {\em Introduction to probability, statistics, and random processes}.
\newblock Kappa Research, LLC Blue Bell, PA, USA, 2014.
\newblock Section 3.2.1: Cumulative Distribution Function.

\bibitem{cai2021ikd}
Y.~Cai, W.~Xu, and F.~Zhang, ``ikd-tree: An incremental kd tree for robotic applications,'' {\em arXiv preprint arXiv:2102.10808}, 2021.

\bibitem{liu1989limited}
D.~C. Liu and J.~Nocedal, ``On the limited memory bfgs method for large scale optimization,'' {\em Mathematical programming}, vol.~45, no.~1, pp.~503--528, 1989.

\bibitem{chen2021clustering}
J.~Chen, C.~Du, Y.~Zhang, P.~Han, and W.~Wei, ``A clustering-based coverage path planning method for autonomous heterogeneous uavs,'' {\em IEEE Transactions on Intelligent Transportation Systems}, vol.~23, no.~12, pp.~25546--25556, 2021.

\bibitem{ni2024cooperative}
J.~Ni, Y.~Gu, G.~Tang, C.~Ke, and Y.~Gu, ``Cooperative coverage path planning for multi-mobile robots based on improved k-means clustering and deep reinforcement learning,'' {\em Electronics}, vol.~13, no.~5, p.~944, 2024.

\bibitem{EurekaIntro}
Eureka, ``Eureka j15 pro ultra flagship robot vacuum,'' 2025.

\bibitem{zhou2019robust}
B.~Zhou, F.~Gao, L.~Wang, C.~Liu, and S.~Shen, ``Fast-planner: An efficient coverage path planning framework for unmanned aerial vehicles,'' {\em IEEE Robotics and Automation Letters}, vol.~8, no.~4, pp.~3529--3536, 2019.

\bibitem{usenko2017real}
V.~Usenko, L.~Von~Stumberg, A.~Pangercic, and D.~Cremers, ``Real-time trajectory replanning for mavs using uniform b-splines and a 3d circular buffer,'' in {\em 2017 IEEE/RSJ International Conference on Intelligent Robots and Systems (IROS)}, pp.~215--222, IEEE, 2017.

\end{thebibliography}

\end{document}